\documentclass[final]{AeroBest_paper}

\usepackage{graphicx}
\usepackage{ragged2e}

\usepackage{subcaption}

\usepackage{amsmath}
\usepackage{amsfonts}

\usepackage{hyperref} 
\hypersetup{pdfborder=0 0 0}

\usepackage{doi}

\usepackage[square,numbers,sort&compress]{natbib}
\usepackage{bm}
\usepackage{comment}
\usepackage{mathrsfs}   

\usepackage[ruled,vlined]{algorithm2e}  
\usepackage{changepage}   

\usepackage[acronyms,nonumberlist,toc,nopostdot]{glossaries}
\makeglossaries

\newacronym{ac:onera}{ONERA}{Office National d’Etudes et de Recherches Aérospatiales}
\newacronym{ac:adoe}{ADoE}{Adaptive Design of Experiments}
\newacronym{ac:doe}{DoE}{Design of Experiments}
\newacronym{ac:gp}{GP}{Gaussian Process}
\newacronym{ac:gpr}{GPR}{Gaussian Process Regression}
\newacronym{ac:mfk}{MFK}{Multi-fidelity Kriging}
\newacronym{ac:mfgp}{MFGP}{Multi-fidelity Gaussian Process}
\newacronym{ac:lhs}{LHS}{Latin Hypercube Sampling}
\newacronym{ac:fitc}{FITC}{Fully Independent Training Conditional}
\newacronym{ac:vfe}{VFE}{Variational Free Energy}
\newacronym{ac:svgp}{SVGP}{Sparse and Variational Gaussian Process}
\newacronym{ac:kl}{KL}{Kullback-Leibler}
\newacronym{ac:elbo}{ELBO}{Evidence Lower BOund}
\newacronym{ac:mmse}{MMSE}{Maximum Mean Square Error}
\newacronym{ac:imse}{IMSE}{Integrated Mean Square Error}
\newacronym{ac:vmr}{VMR}{Variance to Mean Ratio}
\newacronym{ac:ivmr}{IVMR}{Integrated Variance to Mean Ratio}
\newacronym{ac:hf}{HF}{High-fidelity}
\newacronym{ac:lf}{LF}{Low-fidelity}
\newacronym{ac:wt}{WT}{Wind-Tunnel}
\newacronym{ac:cfd}{CFD}{Computational Fluid Dynamics}
\newacronym{ac:slsqp}{SLSQP}{Sequential Least Squares Programming}
\newacronym{ac:segomoe}{SEGOMOE}{Super Efficient Global Optimization with Mixture of Experts}
\newacronym{ac:smt}{SMT}{Surrogate Modeling Toolbox}
\newacronym{ac:dtis}{DTIS}{Département Traitement de l’Information et Systèmes}
\newacronym{ac:m2ci}{M2CI}{Méthodes Multidisciplinaires, Concepts Intégrés}
\newacronym{ac:blup}{BLUP}{Best Linear Unbiaised Prediction}
\newacronym{ac:se}{SE}{Squared Exponential}
\newacronym{ac:rbf}{RBF}{Radial Basis Function}
\newacronym{ac:cdf}{CDF}{Cumulative Distribution Function}
\newacronym{ac:rmse}{RMSE}{Root Mean Square Error}
\newacronym{ac:isae}{ISAE}{Institut Supérieur de l'Aéronautique et de l'Espace}
\newacronym{ac:cobyla}{COBYLA}{Constrained Optimization By Linear Approximation}

\usepackage{afterpage}

\newcommand{\R}{\mathbb{R}}

\newcommand{\Dset}{\mathcal{D}} 

\newcommand{\xbf}{\mathbf{x}}
\newcommand{\ybf}{\mathbf{y}}
\newcommand{\fbf}{\mathbf{f}}
\newcommand{\mbf}{\mathbf{m}}
\newcommand{\Kbf}{\mathbf{K}}
\newcommand{\Xbf}{\mathbf{X}}

\newcommand{\Btheta}{\boldsymbol{\theta}}

\newcommand{\rn}{\mathbf{r}_n}
\newcommand{\BR}{\mathbf{R}}

\usepackage{amssymb}

\usepackage{float}
\usepackage{rotating}

\title{Frequency-Aware Surrogate Modeling with SMT Kernels for Advanced Data Forecasting}

\author{
N. Gonel$^{\bf {1,2*}}$, P. Saves$^{\bf {3}}$, J. Morlier$^{\bf {4}}$}
\heading{N. Gonel, P. Saves, J. Morlier}
\address{
1: Fédération ENAC ISAE-SUPAERO ONERA,\\ Université de Toulouse, \\ 31000, Toulouse, France. \\
\
\\
2: IFP Energies nouvelles, \\  Institut Carnot IFPEN Transports Energie, \\ 92852  Rueil-Malmaison,  France. \\
nicolas.gonel@ifpen.fr\\
\
\\
3:  IRIT,  \\ Université de Toulouse, CNRS, Toulouse INP, UT3,  UT Capitole, \\  Toulouse, France. \\    
 paul.saves@irit.fr \\
 \
\\
4: ICA, \\ 
ISAE--SUPAERO, INSA, CNRS, MINES ALBI, UPS,\\
Université de Toulouse, \\ 31400 Toulouse, France. \\    
joseph.morlier@isae-supaero.fr \\
}

\abstract{
This paper introduces a comprehensive open-source framework for developing correlation kernels, with a particular focus on user-defined and composition of kernels for surrogate modeling.  By advancing kernel-based modeling techniques, we incorporate frequency-aware elements that effectively capture complex mechanical behaviors and time-frequency dynamics intrinsic to aircraft systems. Traditional kernel functions, often limited to exponential-based methods, are extended to include a wider range of kernels such as exponential squared sine and rational quadratic kernels, along with their respective first- and second-order derivatives. The proposed methodologies are first validated on a sinus cardinal test case and then applied to forecasting Mauna-Loa Carbon Dioxide ($\text{CO}_2$) concentrations and airline passenger traffic. 
All these advancements are integrated into the open-source Surrogate Modeling Toolbox (SMT 2.0), providing a versatile platform for both standard and customizable kernel configurations. Furthermore, the framework enables the combination of various kernels to leverage their unique strengths into composite models tailored to specific problems. The resulting framework offers a flexible toolset for engineers and researchers, paving the way for numerous future applications in metamodeling for complex, frequency-sensitive domains.

}


\begin{document}

\section{Introduction}
Over the past few decades, the aeronautical industry has witnessed significant advancements driven primarily by incremental improvements in aircraft design optimization~\cite{chan2022trying}. The design process for aeronautic systems is inherently complex and multidisciplinary, requiring a careful balance between various domains such as aerodynamics, structural integrity, geometry, performance, weight management, and dynamic stability~\cite{SEGONacelle}. This integrated process, referred to as Multidisciplinary Design Analysis (MDA), lays the foundation for the iterative improvements carried out in Multidisciplinary Design Optimization (MDO)~\cite{Lambe2011, Gray}. In our approach to MDO, we adopt the multidisciplinary design feasible strategy, which ensures that the equilibrium between disciplines is re-evaluated and maintained, through a fixed point algorithm, at every modification of the design variables during optimization. However, this approach significantly increases computational demands, posing a major challenge.

To address these challenges, research has increasingly turned to surrogate modeling techniques, which offer an efficient alternative to high-fidelity simulations~\cite{saves2022general, hernandez2010stochastic, costa2016exploratory}.Sofware such as the Surrogate Modeling Toolbox (SMT) have been developed to reduce computational costs by leveraging Gaussian Process (GP) interpolation and Bayesian Optimization (BO) methods~\cite{saves2023smt, saves2023mixed}. However, traditional kernel functions used in GP models—primarily based on exponential formulations—often struggle to capture the complex cyclic and high-dimensional phenomena encountered in advanced aerospace applications~\cite{priem2025high}.

This paper introduces a comprehensive open-source framework for the development of advanced correlation kernels, specifically designed for surrogate modeling in frequency-sensitive environments. Our framework extends conventional methods by incorporating frequency-aware elements into the kernel design. In doing so, we broaden the list of available kernels to include not only standard forms but also alternatives such as exponential squared sine and rational quadratic kernels, along with their first and second-order derivatives. These enhancements enable the efficient optimization of surrogate models that address intricate mechanical behaviors and time-frequency dynamics, which are crucial for analyzing vibrational and aerodynamic characteristics of aircraft structures.

The proposed methodologies are validated through a series of test cases, beginning with a sinus cardinal example and extending to real-world applications such as forecasting Mauna-Loa CO$_2$ concentrations and airline passenger traffic.
All these advancements are encapsulated within SMT 2.0, offering a versatile platform that supports both standard and user-defined kernel configurations. By allowing the combination of multiple kernel types, our approach enables the construction of composite models that leverage the unique strengths of each kernel to address specific modeling challenges.

This extended framework not only enhances predictive capabilities in the preliminary design phase of aerospace engineering but also paves the way for future applications in diverse fields where complex, frequency-sensitive phenomena are of interest. The remainder of the paper is organized as follows: Section~\ref{sec:gp} reviews the background of surrogate modeling and kernel methods in aerospace applications; Section~\ref{sec:dev}
 details the theoretical underpinnings and implementation of our advanced kernel framework; Section~\ref{sec:res} presents validation results on benchmark cases and real-world applications; and Section~\ref{sec:conclu} concludes with a discussion of future directions and potential extensions.

\section{Gaussian processes}
\label{sec:gp}
A \gls{ac:gp} is a stochastic process, \textit{i.e.} a collection of random variables, where any finite collection of those variables has a joint Gaussian distribution. Hereinafter, we detail the mathematical aspects of GP interpolation, also known as Kriging~\cite{krige1951statistical}. 
First, the GP distribution is completely determined by its mean and covariance functions~\cite{williams2006Gaussian}. 
Formally, let $f$ denote the (scalar) target function we wish to estimate:
\begin{equation}
    \begin{matrix}
    f & : & \Dset\subseteq \R^d & \rightarrow & \R\\
      &   &  \xbf & \mapsto & f(\xbf)
    \end{matrix}
\end{equation}
where $\xbf$ is a vector of input variables from the input set $\Dset$ called the \textit{design space}.
In the function-space view, we assume that $f(\xbf)$ is a realization of a GP $Y$ with mean function $m$ and covariance function $k$:
\begin{equation}
    Y \sim \mathcal{GP}(m,k)
\end{equation}
Indeed, the \gls{ac:gp} defines a probability distribution (called the \textit{prior}) over possible functions that fit the set of observations. The main advantage of the GP modeling is that any finite set of $n$ input points $\Xbf=\{\xbf_1,\xbf_2,...,\xbf_n\}\subset\Dset$, the corresponding observations follow a multivariate Gaussian distribution:
\begin{equation}
    \underbrace{
    \begin{bmatrix}
    f(\xbf_1) \\ f(\xbf_2) \\ \vdots \\ f(\xbf_n)
    \end{bmatrix}
    }_\textrm{outputs vector $\fbf$}
    \sim
    \mathcal{N}
    \left(\vphantom{\begin{bmatrix}
        m(\xbf_1) \\ m(\xbf_2) \\ \vdots \\ m(\xbf_n)
    \end{bmatrix}}\right.
    \underbrace{
    \begin{bmatrix}
        m(\xbf_1) \\ m(\xbf_2) \\ \vdots \\ m(\xbf_n)
    \end{bmatrix}
    }_\textrm{mean vector $\mbf$},
    \underbrace{
    \begin{bmatrix}
        k(\xbf_1,\xbf_1) & k(\xbf_1,\xbf_2) & \cdots & k(\xbf_1,\xbf_n) \\
        k(\xbf_1,\xbf_2)^T & k(\xbf_2,\xbf_2) & \cdots & k(\xbf_2,\xbf_n) \\
        \vdots & \vdots & \ddots & \vdots \\
        k(\xbf_1,\xbf_n)^T & k(\xbf_2,\xbf_n)^T & \cdots & k(\xbf_n,\xbf_n)
    \end{bmatrix}}_\textrm{covariance matrix $\Kbf$}
    \left.\vphantom{\begin{bmatrix}
        m(\xbf_1) \\ m(\xbf_2) \\ \vdots \\ m(\xbf_n)
    \end{bmatrix}}\right).
\end{equation}
We can also write it more compact using the shorthands {$\fbf=f(\Xbf)$, $\mbf=m(\Xbf)$ and $\Kbf=k(\Xbf,\Xbf)$}:
\begin{equation}
    p(f(\xbf_1),...,f(\xbf_N)) = p(\fbf\;|\;\Xbf)=\mathcal{N}(\mbf,\Kbf).
\end{equation}
We highlight the fact that the covariance matrix $\Kbf$ (also called \textit{Gram matrix}) is a Positive Semi-Definite (PSD) and symmetric $n\times n$ matrix~\cite{Lee2011} whose components are given by the covariance function $k$: $K_{i,j}=k(\xbf_i,\xbf_j)$.
An example of a valid (stationary) covariance function is the anisotropic\footnote{In the isotropic version, $\theta$ is scalar so the length-scale is the same for each dimension.} \gls{ac:se} kernel\footnote{This kernel is also denoted as the Radial Basis Function (RBF) kernel}:
\begin{equation}
    K^{\theta_i}(\xbf_i,\xbf_j)=\sigma^2\prod\limits_{i=1}^n\exp\left(-\theta_i\left(\xbf_i-\xbf_j\right)^2\right).
\end{equation}
where $\sigma^2$ is the process variance parameter and $\Btheta=\{\theta_1,...,\theta_n\}$ is the vector of inverse length-scale parameters. We call these quantities the hyperparameters of the \gls{ac:gp} that are needed to be estimated in order to best fit the model to our target function.

To make predictions $\fbf_*$ for a new set of $n_*$ inputs $\Xbf_*$ (independent of $\Xbf$), we first write the joint distribution of $\fbf$ and $\fbf_*$ which also follows a multivariate Gaussian distribution:
\begin{equation}\label{eq:joint_dist1}
    \begin{bmatrix}
        \fbf \\ \fbf_*
    \end{bmatrix}
    \sim
    \mathcal{N}\left(
    \begin{bmatrix}
        \mbf \\ \mbf_*
    \end{bmatrix}\;,\;
    \begin{bmatrix}
        \Kbf & \Kbf_* \\
        \Kbf_*^\top & \Kbf_{**}
    \end{bmatrix}
    \right).
\end{equation}
where $\Kbf_*=k(\Xbf,\Xbf_*)$ denotes the $n\times n_*$ matrix of covariance between all pairs of training and test points, and similarly $\Kbf_{**}=k(\Xbf_*,\Xbf_*)$ is the covariance matrix of the test set.

In practice, the true function value value $f(\xbf)$ is often not directly observable. Instead, we observe a noisy output $y(\xbf)=f(\xbf) + \varepsilon$. We assume that this additive noise is an independent and identically distributed (i.i.d.) Gaussian noise~\cite{condearenzana} and we denote its variance $\eta^2$. Using the previous shorthands and letting $\ybf$ denotes the vector of noisy observations at inputs $\Xbf$, the joint distribution \eqref{eq:joint_dist1} then becomes:
\begin{equation}\label{eq:joint_dist2}
    \begin{bmatrix}
        \ybf \\ \fbf_*
    \end{bmatrix}
    \sim
    \mathcal{N}\left(
    \begin{bmatrix}
        \mbf \\ \mbf_*
    \end{bmatrix}\;,\;
    \begin{bmatrix}
        \Kbf + \eta^2\textbf{I} & \Kbf_* \\
        \Kbf_*^\top & \Kbf_{**}
    \end{bmatrix}
    \right).
\end{equation}
Thus, we compute the \textit{posterior} distribution over functions by conditioning the joint prior \eqref{eq:joint_dist2} with the observed data. Using the properties of a multivariate normal distribution, we obtain the predicted distribution:
\begin{equation}
    \fbf_*\;|\;\ybf,\Xbf,\Xbf_* \sim \mathcal{N}\left(\mbf_* + \Kbf_*^\top\left[\Kbf+\eta^2\textbf{I}\right]^{-1}(\ybf-\mbf),\Kbf_{**}-\Kbf^\top\left[\Kbf+\eta^2\textbf{I}\right]^{-1}\Kbf_*\right).
\end{equation}
For the ease of notations, we will then only consider centered \gls{ac:gp}s, \textit{i.e.} zero mean functions: $\mbf=\mbf_*=\boldsymbol{0}$, but equations can be generalized for more general cases with constant (simple Kriging), linear (ordinary Kriging) or polynomial (universal Kriging) priors, adding hyperparameters to the model \cite{williams2006Gaussian}. \\
Using the usual formalism of \gls{ac:gp}, we can rewrite the conditional distribution $\fbf_*\;|\;\ybf$ as:
\begin{equation}
    Z_n \triangleq \Tilde{Y}\left|\left\{\Tilde{Y}(\xbf_i)=y(\xbf_i)\;,\;1\leq i \leq n\right\}\right. \sim \mathcal{GP}(\mu_n,c_n).
\end{equation}
where the posterior mean and covariance functions are defined as follows:
\begin{align}
    \mu_n(\xbf) &\triangleq k(\xbf,\Xbf)\left[\Kbf^{-1} + \eta^2\textbf{I}\right]\ybf \label{eq:mean_func_gp}. \\
    c_n(\xbf,\xbf') &\triangleq k(\xbf',\xbf') - k(\xbf',\Xbf)\left[\Kbf + \eta^2\textbf{I}\right]^{-1}k(\Xbf,\xbf'). \label{eq:covar_func_gp}
\end{align}
An alternative way of formulating the above functions, implemented in  \gls{ac:smt}~\cite{SMT2019,saves2023smt}, is to introduce the correlation function $R$ and to consider a unique \gls{ac:gp} variance $\sigma^2$ that is an homoscedastic process:
\begin{equation}
    \text{Cov}\left(Y(\xbf),Y(\xbf')\right) = k(\xbf,\xbf') = \sigma^2 R(\xbf,\xbf')\label{eq:cond_proc}.
\end{equation}
%
With this new notation, the equations (\ref{eq:mean_func_gp}) and (\ref{eq:covar_func_gp}) can be rewritten as follows~\cite{jones1998efficient}:
\begin{align}
    \mu_n(\xbf) &= \rn^\top(\xbf)\BR_n^{-1}\ybf \label{eq:mean_func_gp2}. \\
    c_n(\xbf,\xbf') &= \sigma^2\left[r(\xbf,\xbf') - \rn^\top(\xbf)\BR_n^{-1}\rn(\xbf')\right]. \label{eq:covar_func_gp2}
\end{align}
where $\ybf = \left[y(\xbf_i),...,y(\xbf_n)\right]^\top$ is the observations vector, $\rn(\xbf)=k(\Xbf,\xbf)=\left(k(\xbf_i,\xbf)\right)_{1\leq i \leq n}$ is the cross-covariance vector and $\BR_n=\Kbf+\eta^2\textbf{I}=\left(k(\xbf_i,\xbf_j)\right)_{1\leq i,j\leq n}$ is the covariance matrix
Note that if we consider noise-free observations (\textit{i.e.} $\eta^2=0$), we simply have $\BR_n=\Kbf$. The conditional process defined by Eq.(\ref{eq:cond_proc}) is then used as a surrogate model to approximate our target function $f$.

To conclude this short presentation of the GP regression model, we address the optimization of the hyperparameters mentioned above. Indeed, it is necessary to tune the model parameters $\sigma^2,\Btheta$ and $\eta^2$ on the training set $\Xbf$ before making predictions at new inputs $\Xbf_*$.  
Here, everything being Gaussian and our prior being an uniform distribution between given bounds, we only optimize the \textit{marginal likelihood} $p(\ybf)$ to obtain our optimal hyperparameters~\cite{MLE}. This likelihood is given by: 
\begin{equation}\label{eq:marginal_likelihood}
    p(\ybf) = \mathcal{N}(\boldsymbol{0},\Kbf + \eta^2\textbf{I}) = \frac{1}{(2\pi)^{N/2}\text{det}(\Kbf+\eta^2\textbf{I})^{1/2}}\exp\left(-\frac{1}{2}\ybf^\top\left(\Kbf+\eta^2\textbf{I}\right)\ybf\right).
\end{equation}
The optimal set of hyperparameters is then obtained by maximizing the log marginal likelihood:
\begin{equation}
    \log p(\ybf) = -\frac{1}{2}\log\left(\text{det}\left(\Kbf + \eta^2\textbf{I}\right)\right) - \frac{1}{2}\ybf^\top\left(\Kbf+\eta^2\textbf{I}\right)^{-1}\ybf - \frac{N}{2}\log(2\pi).
\end{equation}
We recall that the marginal likelihood $p(\ybf)$ is the integral of the \textit{likelihood} $p(\ybf|\fbf,\Xbf)$ times the prior $p(\fbf|\Xbf)$. Thus, given the prior $\fbf|\Xbf\sim\mathcal{N}(0,\Kbf)$ and the likelihood $\ybf|\fbf\sim\mathcal{N}(\fbf,\eta^2\textbf{I})$, the computation leads to \eqref{eq:marginal_likelihood}.
We saw that to define the GP, we need to define its kernel first. This kernel will give the structure of the GP. We already showed the squared exponential kernel but other kernels exists in SMT such as the absolute exponential, the Matern kernel or the power exponential kernel.

\section{A Novel User-Defined Kernel Library for Gaussian Process}
\label{sec:dev}

The SMT 2.0 interface is more user-friendly and offers improved and detailed documentation for both users and developers. Hosted publicly, SMT 2.0 can be directly imported within Python scripts, is released under the New BSD license, and runs on Linux, macOS, and Windows operating systems. Furthermore, regression tests are automatically executed on each supported platform whenever changes are committed to the repository, ensuring robustness and compatibility.
To enhance the modeling capabilities of Gaussian Processes (GP), we have expanded SMT by implementing two new kernels:

\begin{itemize}
    \item \textbf{Rational Quadratic Kernel}: This kernel is particularly effective for modeling functions that exhibit irregularities. It can be interpreted as a scale mixture of squared exponential kernels, thereby allowing for varying length scales. Its formula is given by:
    \begin{equation}
        K^{\theta_l,\theta_k}(x, x') = \left( 1 + \frac{(x - x')^2}{\theta_l} \right)^{-\theta_k}
    \end{equation}
    where \(\theta_l\) controls the length scale and \(\theta_k\) adjusts the weighting across scales.
    
    \item \textbf{Periodic Kernel}: Useful for modeling periodic data, this kernel captures cyclic patterns inherent in many physical and time-series processes. Its formula is defined as:
    \begin{equation}
        K^{\theta_l,\theta_k}(x, x') = \exp\left(-\frac{\sin^2((x_l-x_l')\theta_l)}{\theta_k}\right)
    \end{equation}
    where \(\theta_l\) controls the periodicity and \(\theta_k\) determines the smoothness of the periodic oscillations.
\end{itemize}

In order to further expand the range of functions that can be modeled with GP, it is essential to leverage the property of kernel composition. This property states that the sum and product of PSD kernels are also valid kernels. Adding kernels models the superposition of independent functions, while multiplying them accounts for interactions between different phenomena \cite{duvenaud2013structure}. However, SMT previously did not offer a native mechanism for combining kernels, so we introduce a novel architecture that supports flexible kernel composition.

\begin{figure}[H]
    \hspace{-0.5cm}
    \includegraphics[width=1.1\textwidth]{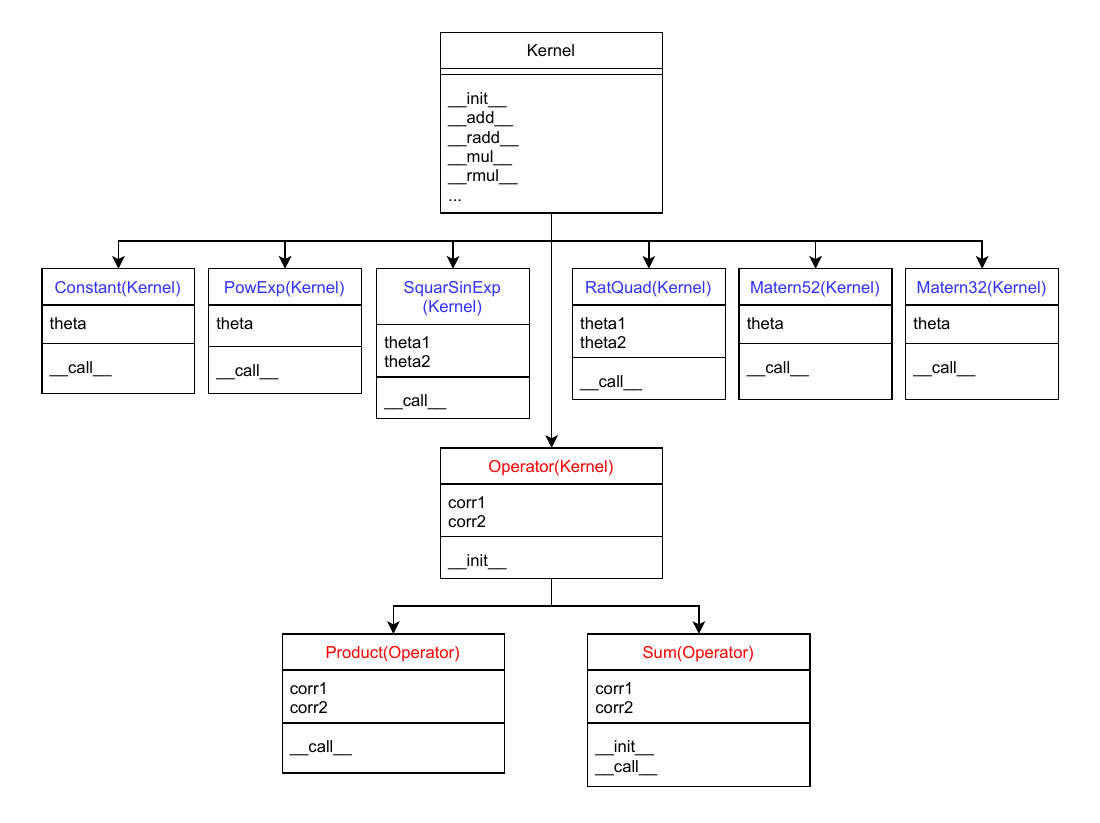} 
    \caption{Class diagram for kernels in SMT. In black, the main class, in blue the basic kernels already implemented, and in red the operator.}
        \label{fig:class_diagram}
\end{figure}

The optimal structure for managing a wide variety of kernel functions is a class-based architecture. To address the need for flexibility and ease of use, we have defined a new class structure within SMT that can accommodate any kernel form. The architecture, illustrated in Fig.~\ref{fig:class_diagram}, is composed of three main elements:

\begin{enumerate}
    \item \textbf{Base Kernel Class}: An abstract class that provides a template for all kernel implementations. Any new kernel can inherit from this class to ensure consistency and compatibility.
    
    \item \textbf{Standard Kernel Implementations}: Existing kernels in SMT have been refactored into class-based structures. This includes not only the traditional kernels (such as the squared exponential or Matern kernels) but also the newly added rational quadratic and periodic kernels.
    
    \item \textbf{Kernel Composition Utilities}: To simplify the process of creating complex kernels, we provide built-in tools that allow users to combine kernels using basic Python operators (e.g., addition and multiplication) as the product and sum of SDP kernels still results in a kernel~\cite{roustant2020group}. These utilities enable the effortless formation of composite kernels, which can model both the independent superposition and interactions between different data features.
\end{enumerate}

The new architecture empowers users to construct expressive, hybrid kernels tailored to specific applications. For example, in modeling the oscillatory behavior of an aircraft wing subject to aerodynamic forces, one might combine a periodic kernel to capture cyclic oscillations with a squared exponential kernel to model broader trends.  By integrating these advancements, SMT 2.0 significantly extends the flexibility and power of Gaussian Process modeling. This novel kernel library not only supports standard kernel configurations but also facilitates the development of custom, composite models tailored to the unique challenges of complex, frequency-sensitive applications as shown in the following section.

\section{Results}
\label{sec:res}

To test our framework, we evaluated its ability to model a range of functions and datasets, and assess its capacity to uncover underlying structures and extrapolate beyond the given data. Below, we provide results from different case studies, each with its own characteristics and challenges.

\begin{figure}[h!]
    \centering
    \includegraphics[width=0.8\textwidth]{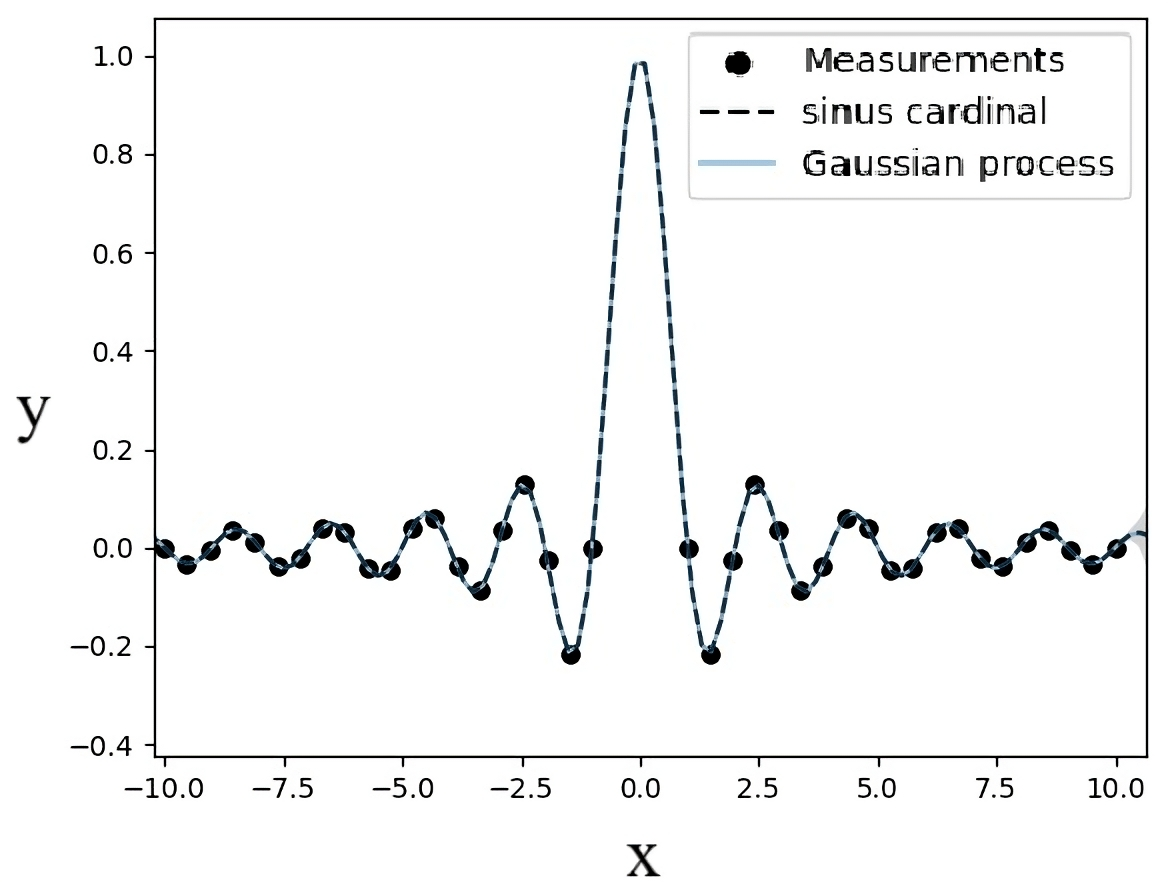}
    \caption{Posterior GP with kernel of Eq.(\ref{sinckernel}) on the sinc function}
    \label{fig:sinc}
\end{figure}

\paragraph{Cardinal Sine function}
We begin by testing our framework with the well-known cardinal sine (sinc) function. This function, although not strictly periodic, exhibits a periodic component with a varying amplitude. The sinc function is a good candidate for testing, as it has a periodicity that is modified by amplitude fluctuations, which we aim to capture with our kernel model. We employed a kernel formed by the multiplication of a squared exponential (SE) kernel and a periodic kernel:
\begin{equation}\label{sinckernel}
\begin{aligned}
K(x,x') = K_{SE}^{\theta_1}(x, x') \cdot K_{Periodic}^{\theta_2, \theta_3}(x, x')
\end{aligned}
\end{equation}
This kernel combination allows us to model both the periodic nature of the function and the amplitude variations. As shown in Fig.~\ref{fig:sinc}, the model perfectly captures the periodic structure and the largest amplitude, providing a good representation of the sinc function over the given input domain.

\paragraph{Mauna-Loa $\text{CO}_2$ observations}
Next, we apply our framework to the Mauna-Loa $\text{CO}_2$ dataset~\cite{keeling1978influence}, which contains historical data on the concentration of atmospheric $\text{CO}_2$ since 1958. The Mauna-Loa dataset is often used to test models for environmental data, as it contains a combination of long-term trends, seasonal oscillations, and noise. The primary goal of this case study is to test our model's ability to capture these components, particularly the periodic oscillations and the irregularities introduced by noise.

\begin{figure}[h!]
    \centering
    \includegraphics[width=0.8\textwidth]{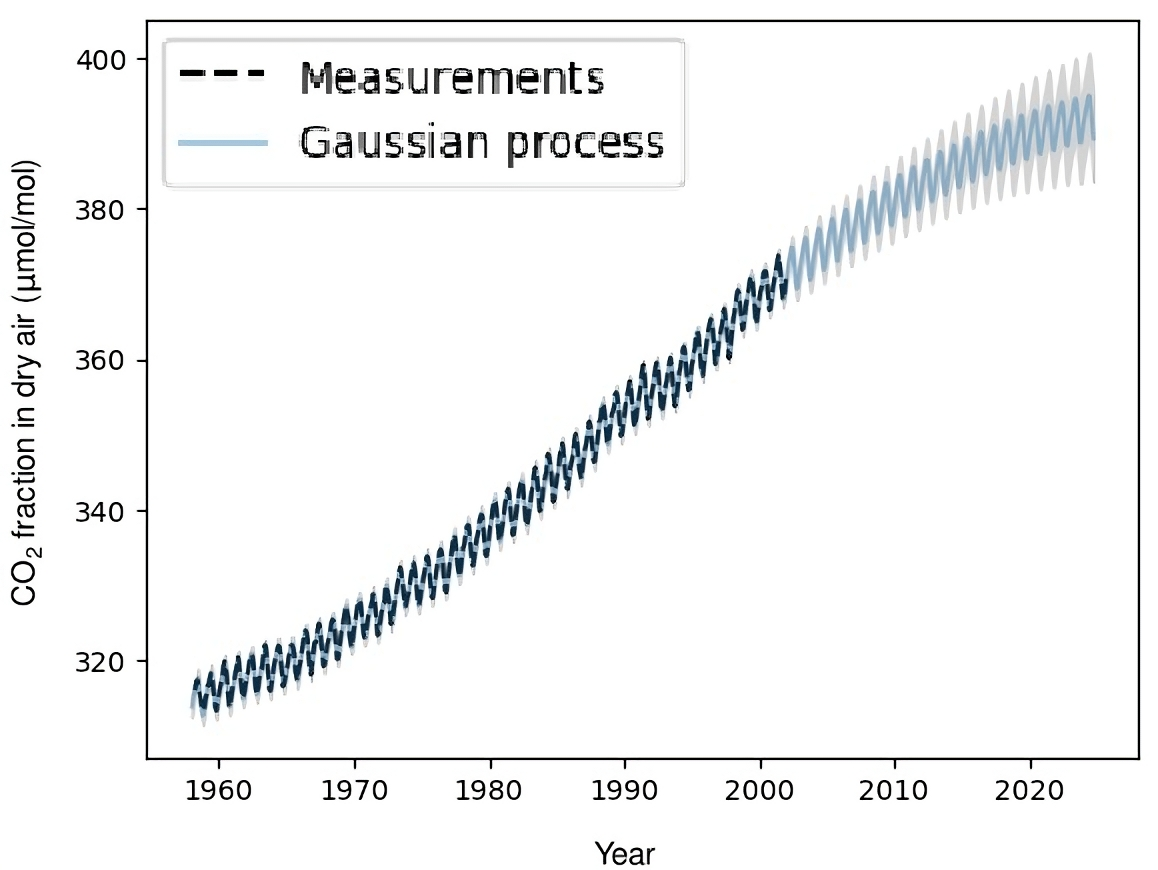} 
    \caption{$\text{CO}_2$ forecasting with GP kernel of Eq.(\ref{4kernel1}).}
    \label{fig:mauna_loa_smt_pred}
\end{figure}

To model this dataset, we employed a kernel that combines multiple kernel functions to address the different behaviors in the data. Specifically, we use a periodic kernel to model the oscillations, a squared exponential kernel to model the small variations around the periodicity, another squared exponential kernel to account for the linear growth trend, and a rational quadratic kernel to handle irregularities:
\begin{equation}\label{4kernel1}
\begin{aligned}
K(x,x') = &\ K_{SE}^{\theta_1}(x, x') \cdot K_{Periodic}^{\theta_2, \theta_3}(x, x') \\
&+ K_{SE}^{\theta_4}(x, x') + K_{rational\_quadratic}^{\theta_5, \theta_6}(x, x')
\end{aligned}
\end{equation}
The results in Fig.~\ref{fig:mauna_loa_smt_pred} show that the model effectively captures the periodicity and growth, with the interpolation fitting the observed data closely. Additionally, the extrapolation provides reasonable forecasts, with the identified components appearing to repeat in the future, as expected from the underlying structure of the data.

\paragraph{Airlines Annual Passengers Traffic}
Another case we examine is a dataset containing monthly international airline passenger counts from January 1949 to December 1960~\cite{box2015time}. This dataset is commonly used for testing forecasting models due to its long-term trend, seasonal fluctuations, and sudden changes in direction. Similar to the Mauna-Loa dataset, this data exhibits three primary components: a long-term trend, periodic oscillations, and sharp shifts in direction, such as a sudden increase in the number of passengers during specific periods.

\begin{figure}[H]
    \centering
    \includegraphics[width=1.0\textwidth]{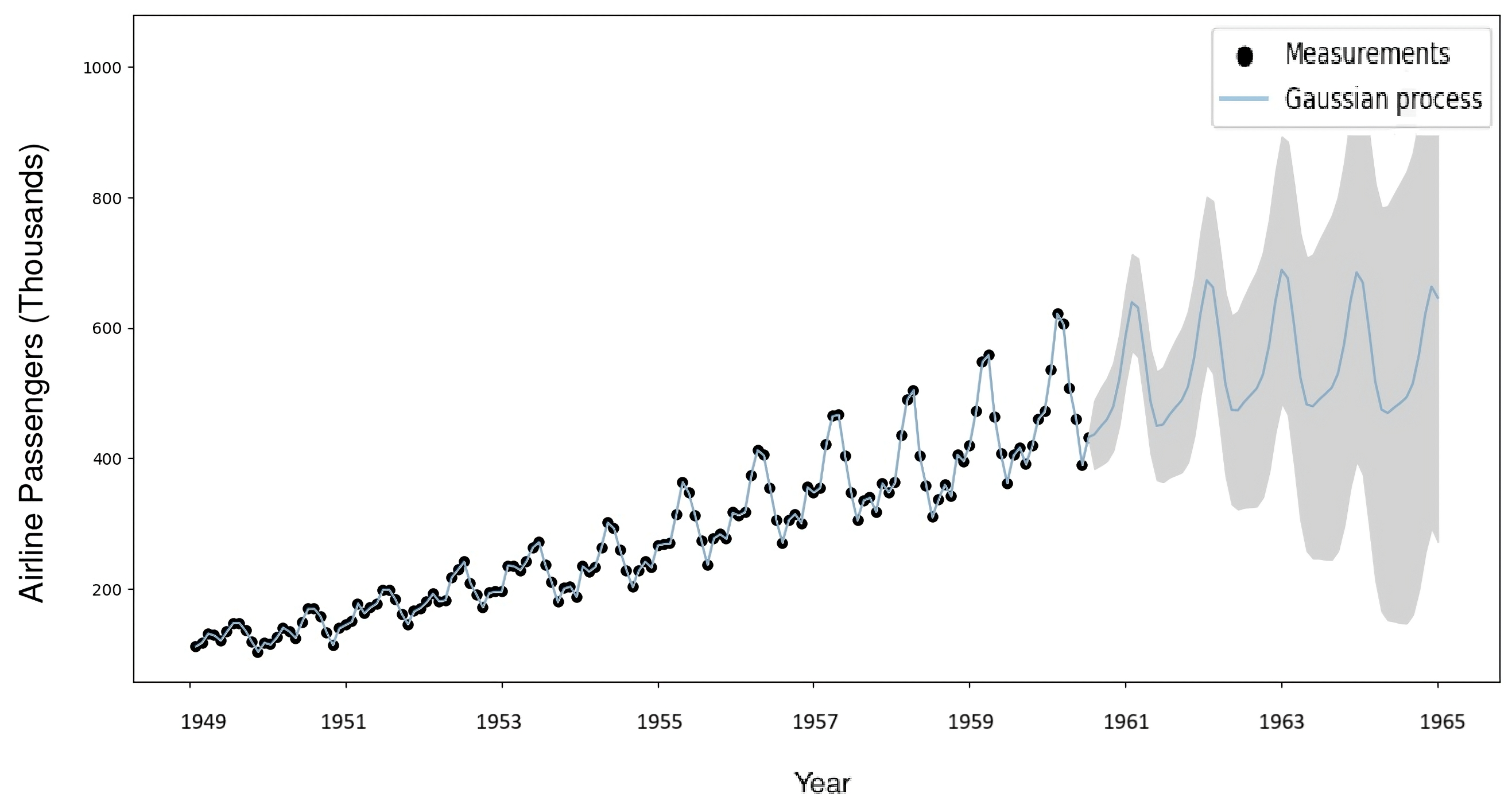} 
    \caption{GP approximation with a kernel given by Eq.(\ref{4kernel2}) on passengers database}
    \label{fig:atm_model}
\end{figure}

To model this dataset, we apply the same kernel we used for the Mauna-Loa data, given its similar structure:
\begin{equation}\label{4kernel2}
\begin{aligned}
K(x,x') = &\ K_{SE}^{\theta_1}(x, x') \cdot K_{Periodic}^{\theta_2, \theta_3}(x, x') \\
&+ K_{SE}^{\theta_4}(x, x') + K_{rational\_quadratic}^{\theta_5, \theta_6}(x, x')
\end{aligned}
\end{equation}
The results shown in Fig.~\ref{fig:atm_model} indicate that the kernel successfully captures the underlying structure of the data. The interpolation is smooth and accurate, and the extrapolated part is plausible, although the variance is higher, possibly reflecting the uncertainty in predicting sudden changes in the future. This suggests that while the model captures the general trends, future data points may have a high degree of uncertainty.

\section{conclusion}
\label{sec:conclu}

In this paper we presented a functional framework to create and combine new kernels for GP in the python library SMT. This expands the scope of functions that can be modeled with this method and also provides a way to use Kriging for prediction and extrapolation with the use of the periodic kernel, which is usually not possible. However, there are still some gray areas. For complex data, finding the right kernel can be a complex endeavor 
and we can note that all the tests done here are in 1 dimension only, as the complexity increases a lot for complex kernel with the dimension. Indeed the number of hyperparameters in the kernel is multiplied by the dimension making the duration of modelisation quickly prohibitive.

The development of this user-defined kernel library opens several avenues for future work.  First this work was limited to sum and product of kernels and can be extended to any PSD matrix manipulation such as ANalysis Of VAriance (ANOVA)~\cite{roustant2020group}. Also, it was limited to spatial classical kernels whereas some more complex ones are better tailored for timeseries forecasting such as the Mercer  or spectral kernels~\cite{Wilson2013, gorodetsky2016mercer}. Future updates may include automated techniques for selecting the most appropriate kernel configuration based on data characteristics, utilizing methods such as Bayesian model selection or evolutionary algorithms.
Enhancing kernel hyperparameter tuning through gradient-based methods and automatic differentiation will further improve computational efficiency, especially for large-scale problems.
The new framework can be directly applied to complex aeroelastic  analyses of high aspect ratio wings, such as HALE (High Altitude, Long Endurance) solar aircraft subjected to gust perturbations.


\section*{Acknowledgements}
This work is part of the activities of ONERA - ISAE - ENAC joint research group.
The research presented in this paper has been performed within the framework of the COLOSSUS project (Collaborative System of Systems xEploration of Aviation Products, Services and Business Models) and has received funding from the European Union Horizon Programme under grant agreement n${^\circ}$101097120. The authors acknowledge the research project MIMICO funded in France by the Agence Nationale de la Recherche (ANR, French National Research Agency), grant number ANR-24-CE23-0380.

\bibliography{main}

\end{document}